%% file: main.tex
\documentclass[10pt,twocolumn,letterpaper]{article}

\usepackage{cvpr}              

\input{preamble}

\definecolor{cvprblue}{rgb}{0.21,0.49,0.74}
\usepackage{bm}
\usepackage{verbatim}
\usepackage{amsmath}
\usepackage{amssymb}
\usepackage{booktabs}
\usepackage{float}
\usepackage{makecell}
\usepackage{xcolor}
\usepackage{pifont}
\newcommand{\cmark}{\ding{51}}%
\newcommand{\xmark}{\ding{55}}%
\usepackage{multicol}
\usepackage{multirow}
\usepackage[pagebackref,breaklinks,colorlinks,citecolor=cvprblue]{hyperref}
\usepackage[accsupp]{axessibility}

\def\paperID{16} 
\def\confName{CVPR}
\def\confYear{2024}

\title{\ours: Token Selective Attention for Efficient Vision Transformers}

\author{
Manish Kumar Singh~~~
Rajeev Yasarla~~~
Hong Cai~~~
Mingu Lee~~~
Fatih Porikli~~~
\smallskip
\\
Qualcomm AI Research$^*$
\\
\smallskip
{\tt\small\{masi, ryasarla, hongcai, mingul, fporikli\}@qti.qualcomm.com}
}

\newcommand{\ours}{{ToSA}\xspace}

\begin{document}
\maketitle
\def\thefootnote{*}\footnotetext{Qualcomm AI Research is an initiative of Qualcomm Technologies, Inc.}
\input{sec/0_abstract}    
\input{sec/1_intro}
\input{sec/3_method}

\input{sec/4_experiments}

\input{sec/5_conclusion}

\newpage
{
    \small
    \bibliographystyle{ieeenat_fullname}
    \bibliography{main}
}


\end{document}

%% file: preamble.tex
\usepackage[dvipsnames]{xcolor}


%% file: sec/0_abstract.tex
\begin{abstract}
\vspace{-5pt}

In this paper, we propose a novel token selective attention approach, \ours, which can identify tokens that need to be attended as well as those that can skip a transformer layer. More specifically, a token selector parses the current attention maps and predicts the attention maps for the next layer, which are then used to select the important tokens that should participate in the attention operation. The remaining tokens simply bypass the next layer and are concatenated with the attended ones to re-form a complete set of tokens. In this way, we reduce the quadratic computation and memory costs as fewer tokens participate in self-attention while maintaining the features for all the image patches throughout the network, which allows it to be used for dense prediction tasks. 
Our experiments show that by applying \ours, we can significantly reduce computation costs while maintaining accuracy on the ImageNet classification benchmark. Furthermore, we evaluate on the dense prediction task of monocular depth estimation on NYU Depth V2, and show that we can achieve similar depth prediction accuracy using a considerably lighter backbone with \ours.


\end{abstract}

%% file: sec/1_intro.tex
\vspace{-10pt}
\section{Introduction}
\label{sec:intro}
\vspace{-3pt}


Vision transformers (ViTs)~\cite{dosovitskiy2020image} have been the core of many latest advances in computer vision, with self-attention playing a critical role to generate key visual features. However, the self-attention operation incurs quadratic computation and memory costs w.r.t. the input size. This makes it expensive and challenging to run vision transformers on high-resolution images and on resource-constrained devices.

\begin{figure}[t!]
    \centering
    \includegraphics[width=0.38\textwidth]{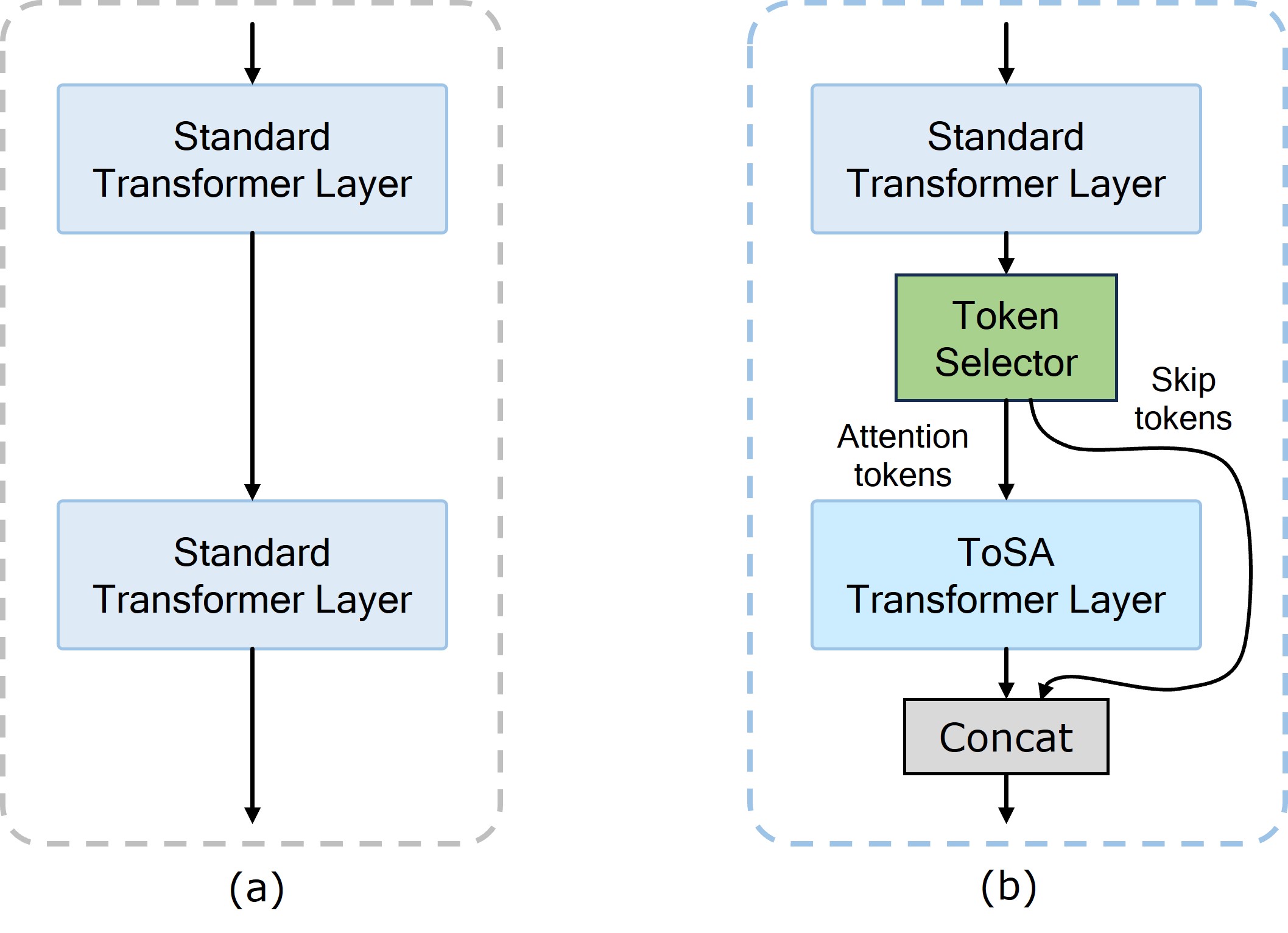}
    \vspace{-8pt}
    \caption{\small (a) Two consecutive standard transformer layers. (b) A standard transformer layer followed by our \ours layer. Our proposed approach operates on a pair of consecutive transformer layers and modifies the second layer to be token selective. In other words, only a subset of tokens participate in the self-attention in the second layer while the rest bypass the layer; our token selector predicts importance scores of the tokens for selection. Note that we retain all the tokens throughout the layers to facilitate dense prediction tasks, e.g., segmentation, depth estimation.}
    \label{fig:teaser}
    \vspace{-10pt}
\end{figure}

Researchers have looked into alternative ways to apply attention to make vision transformers more efficient. For instance, Swin~\cite{liu2021swin} applies self-attention within local windows of the image. Others propose to first apply a few stages of convolutional layers and only apply self-attention on significantly downsampled versions of the input image to reduce computation costs (e.g.,~\cite{li2022efficientformer, vasu2023fastvit, hatamizadeh2023fastervit}). Some other papers design alternative ways to generate attention, e.g., removing softmax~\cite{koohpayegani2024sima}, using ReLU for normalization~\cite{cai2024efficientvit}, applying attention on feature channel dimension~\cite{maaz2022edgenext}, additive attention~\cite{shaker2023swiftformer}, etc.

However, many latest advances in computer vision still predominantly rely on the vanilla ViTs, thanks to several favorable properties: their design is straightforward and easy to implement, there exist powerful pretraining methods and pretrained checkpoints for them (e.g., CLIP~\cite{radford2021learning}, DINOV2~\cite{oquab2023dinov2}), they scale better with massive data~\cite{dehghani2023scaling}, and they may have more consistent performance across problem domains given the generic design of standard self-attention. 

As such, researchers have looked into reducing the computation costs of ViTs while preserving the standard self-attention. A common approach is to reduce the tokens as they go through the network~\cite{bolya2022tome, zhangsynergistic, wang2022vtc, fayyaz2022adaptive}. This effectively reduces the computation and memory usage which scale w.r.t. the number of attended tokens. However, these methods only work for classification tasks. As some of the tokens are discarded or merged during inference, these networks cannot be used for dense prediction tasks which require distinct features for all the image pixels/patches.

In this paper, we propose a novel token selective attention approach, \ours, which can make any vision transformer more efficient by selecting only subsets of tokens for self-attention. More specifically, \ours is applied to two consecutive transformer layers where the latter will become token selective. Given the multi-head attention maps from the first layer, our token selector predicts the attention maps for the next layer and produces important scores for the tokens accordingly. The top important tokens will go into self-attention of the next layer, which we call a \emph{\ours transformer layer}; the \ours transformer layer replaces the original second standard transformer layer. The rest of the tokens simply bypass the \ours layer and are re-joined with the attended tokens. In this way, we reduce the quadratic computation and memory costs as fewer tokens are attended, maintain the favorable standard self-attention operation, and retain the full set of tokens during inference so the network can be used as an encoder for dense prediction tasks. Fig.~\ref{fig:teaser} provides a high-level illustration of \ours.

\begin{figure*}
    \centering
    \includegraphics[width=0.97\textwidth]{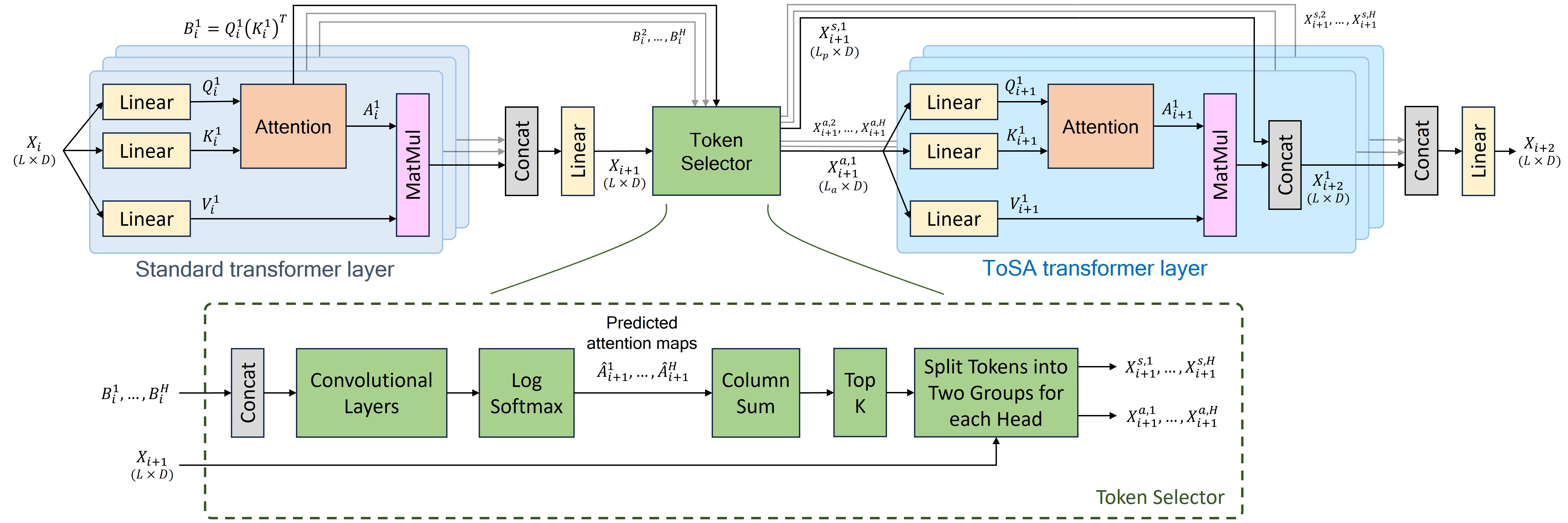}
    \vspace{-5pt}
    \caption{\small Overview of our proposed \ours approach. Based on an input $X_i$, a standard transformer layer generates an output $X_{i+1}$, as well as $QK^T$ maps (prior to Softmax), $B_i^1,B_i^2,...,B_i^H$, for the $H$ attention heads. Based on them, our token selector predicts the attention maps at the next layer and uses them to identify tokens that need to be attended, $X_{i+1}^{a,h}$, and tokens that can skip the next layer, $X_{i+1}^{p,h}$, for each head. Then, self-attention is performed on $X_{i+1}^{a,h}$, the output of which is concatenated with $X_{i+1}^{p,h}$ to form the output of this head, $X_{i+2}^h$. All $X_{i+2}^h$'s are combined to generate the final output of this layer, $X_{i+2}$. By reducing the number of tokens participating in attention, we significantly save computation and memory given the quadratic complexity of self-attention, while slightly improving accuracy.}
    \label{fig:diagram}
    \vspace{-3pt}
\end{figure*}

\begin{figure*}[t!]
    \centering
    \includegraphics[width=0.9\textwidth]{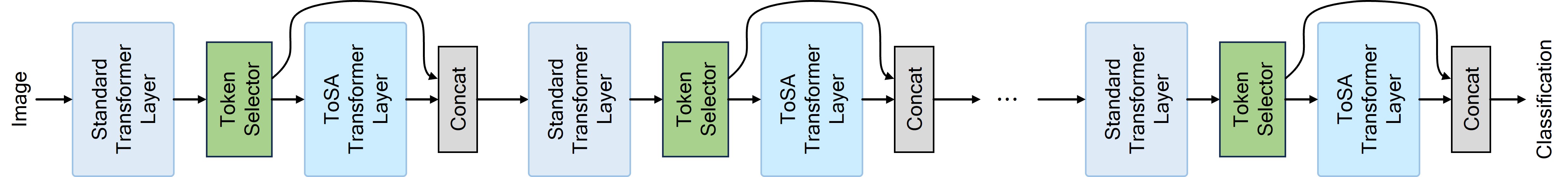}
    \vspace{-5pt}
    \caption{\small Applying our proposed \ours to all the pairs of transformer layers in a 12-layer DeiT.}
    \label{fig:full_deit}
    \vspace{-10pt}
\end{figure*}

In summary, our main contributions are as follows:
\begin{itemize}
    \item We present \ours, a novel approach that improves the efficiency of a multi-layer vision transformer by reducing the number of tokens participating in the attention computation in certain layers. Although not all tokens are attended in \ours, they are retained throughout the layers. This allows the model to be used for dense prediction tasks.
    \item In order to select a subset of tokens to be fed into self-attention, we devise a token selector that predicts important scores of the tokens based on the previous transformer layer. The scores are used to identify tokens that need to be attended at the next layer and the rest will simply skip the next layer. 
    \item We evaluate \ours on the standard ImageNet classification benchmark~\cite{russakovsky2015imagenet}, showing that \ours significantly reduces the computation cost while maintaining accuracy. In addition, we apply \ours to a vision transformer backbone and use it as an encoder for a monocular depth estimation task. We show that our more efficient backbone maintains the depth prediction accuracy.
\end{itemize}

%% file: sec/3_method.tex


\section{Proposed Approach: \ours}
\label{sec:method}
\vspace{-3pt}
In this section, we present our proposed token selective attention method, \ours. \ours selects only a subset of tokens to participate in the self-attention of a transformer layer, which considerably reduces computation and memory costs, while maintaining the full set of image features throughout the network, making it feasible to be used as an encoder for dense prediction tasks. 

\subsection{Self-Attention with Selected Tokens}
\vspace{-3pt}
\textbf{Standard Self-Attention:} Consider an input $X_i \in \mathbb{R}^{L \times D}$, where $L$ is the number of tokens (i.e., features of image patches) and $D$ is the feature dimension; the batch dimension is omitted here for conciseness. In a standard vision transformer layer $i$, $X_i$ first goes through linear layers to generate the query $Q_i^h$, key $K_i^h$, and value $V_i^h$, respectively, for each attention head, $h\in \{1,...,H\}$, as follows: \vspace{-3pt}
\begin{equation}\label{QKV}
\begin{split}
    {Q_i^h} = {W_{Q,i}^h}{X_i}, {K_i^h} = {W_{K,i}^h}{X_i}, {V_i^h} = {W_{V,i}^h}{X_i}, \\[-3pt]
\end{split}
\end{equation}
where $W_{Q,i}^h$, $W_{K,i}^h$, $W_{V,i}^h$ are linear transformation matrices.

Then, an attention map is computed based on the query and key, which is then multiplied with the value: \vspace{-3pt}
\begin{equation}\label{SA}
\begin{split}
    A_i^h = \text{softmax}(\frac{Q_i^h(K_i^h)^T}{\sqrt{D}}),\quad
    X_{i+1}^h = A_i^h \cdot V_i^h. \\[-3pt]
\end{split}
\end{equation}

The outputs from all attention heads are concatenated and fed into a linear layer, $F_i$, to general the final output of this transformer layer: \vspace{-3pt}
\begin{equation}
\begin{split}
    X_{i+1} = F_i(W_{O,i} \cdot \text{Concat}(X_{i+1}^1, ..., X_{i+1}^H)),\\[-3pt]
\end{split}
\end{equation}
where $W_{O,i}$ is a linear projection matrix.

\vspace{5pt}
\noindent \textbf{Token Selective Attention (\ours):} 
In the standard self-attention, all tokens participate in the attention computation, resulting in $\text{O}(N^2)$ computation and memory costs. One way to reduce computation costs is to reduce the number of tokens taking part in the self-attention, i.e., reducing $N$.

In order to do this, we need to identify the subset of important tokens that should go into self-attention, and the remaining set that should skip the attention. We propose a novel token selector, which operates between a standard transformer layer and a selective attention transformer layer (\ours transformer layer). It consumes the un-normalized attention maps (i.e., $QK^T$) from the previous standard layer and predicts the attention maps for the respective heads of the next layer. These predictions are used to identify important tokens. More specifically, we perform column sum over the predicted attention map of each head, which generates an importance score for each token. Based on the important scores, we select the top-K tokens to perform self-attention in the next layer, to which we refer as the \emph{attention tokens} $X_{i+1}^{a,h}$, and K is determined based on a prescribed attention ration $r$. For instance, $r=80\%$ indicates that the top $80\%$ tokens will participate in the self-attention in the next layer. The remaining tokens simply bypass the next transformer layers, which we call \emph{skip tokens} $X_{i+1}^{s,h}$.

After the attention tokens go through the transformer layer, where standard self-attention is applied to this subset, the attended tokens and skipped tokens are combined to re-form the complete set of tokens, for each head, i.e., \vspace{-3pt}
\begin{equation}
\begin{split}
    X_{i+2}^h = \text{Concat}(X_{i+2}^{a,h}, X_{i+1}^{s,h}), \\[-3pt]
\end{split}
\end{equation}
where $X_{i+2}^{a,h}$ is the self-attention output from $X_{i+1}^{a,h}$ and the concatenation is performed along the token dimension. Finally, the outputs from all the heads are concatenated and processed by a linear layer to produce the final output from this \ours layer. Fig.~\ref{fig:diagram} illustrates this process for a pair of standard layer and \ours layer.

\subsection{Token Selector: Architecture and Training}\label{tok_sel_model}
\vspace{-3pt}
A high level overview of the token selector architecture is given in Fig.~\ref{fig:diagram}. 
The token selector takes as input the multi-head un-normalized attention maps ($QK^T$) from the previous standard transformer layer. It consists of two 1D convolutional layers with a intermediate ReLU activation, followed by a log-softmax layer that predicts the full $L\times L$ multi-head attention maps, $\hat{A}_{i+1}^1, ..., \hat{A}_{i+1}^H$, for the next layer; we choose log-softmax for the output layer as it produces normalized attention maps and is more numerically favorable in training.

We train the token selector's attention prediction part based on ground-truth  attention maps computed by the pretrained model. More specifically, given a pre-trained backbone (e.g., DeiT~\cite{touvron2021training} trained on ImageNet classification~\cite{russakovsky2015imagenet}), we insert a token selector between two consecutive standard transformer layers $i$ and $i+1$. During training, the token selector consumes the $QK^T$ maps from the first transformer layer and its predicted attention maps are compared against the actual attention maps computed by the second transformer layer of the pretrained model, which provides supervision to train the token selector, i.e., \vspace{-3pt}
\begin{equation}\label{eq:token selector loss}
\begin{split}
    \mathcal{L}_{i,i+1} = \sum_{h=1}^H \mathcal{L}_\text{KLD}(\hat{A}_{i+1}^h,{A}_{i+1}^h), \\[-3pt]
\end{split}
\end{equation}
where $\mathcal{L}_\text{KLD}$ denotes the KL-Divergence loss. In this process, the pre-trained transformer layers are frozen and only the token selector is trained. Note that we only use the same data used to pre-train the vision transformer model to train our token selectors, and do not use extra training data.

\begin{table*}[t!]
\small
  \centering
  \begin{tabular}{lcccccc}
    \toprule
    \textbf{Method} & Top-1 Acc. & Top-5 Acc. & GFlops & Flops Reduction & Use for Dense Prediction?\\
    \midrule
    DeiT-Tiny~\cite{touvron2021training} & 72.2 & 91.1 & 1.3 & - & \textcolor{Green}{\cmark}\\
    \hline
    ToMe\footnotemark~\cite{bolya2022tome}  & 71.4 & - & 0.8 & 41.7 & \textcolor{red}{\xmark}\\
    VTC-LFC\footnotemark~\cite{wang2022vtc}  & 71.0 & 90.4 & 0.8 & 41.7 & \textcolor{red}{\xmark}\\
    STAR~\cite{zhangsynergistic}  & 72.0 & 90.7 & 0.7 & 46.2 & \textcolor{red}{\xmark}\\
    a-STAR~\cite{zhangsynergistic}  & 72.0 & 90.7 & 0.7 & 46.2 & \textcolor{red}{\xmark}\\
    \hline
    DeiT-Tiny w/ \ours (ours) & 73.9 & 91.9 & 1.0 & 25.8 & \textcolor{Green}{\cmark}\\
    \bottomrule
  \end{tabular}
  \vspace{-7pt}
  \caption{\small Results on ImageNet-1K classification using DeiT-Tiny~\cite{touvron2021training} as the base model. Our proposed \ours significantly reduces the computation cost, while improving accuracy and retaining the full set of tokens. In contrast, other existing methods cause accuracy drops and cannot be used for dense prediction as subsets of tokens (i.e., features of some image patches) are discarded.}
  \label{tab:imagenet}
  \vspace{-8pt}
\end{table*}

\subsection{Full Vision Transformer Model with \ours}
\vspace{-3pt}
Given a pretrained vision transformer model, we can apply \ours to any pair of layers where the second layer will be replaced by a \ours layer and we train a token selector between them. Consider a 12-layer standard vision transformer as an example. We can apply \ours to consecutive pairs of layers, and replace the 2nd, 4th, 6th, 8th, and 10th standard transformer layers with \ours transformer layers; see an illustration in Fig.~\ref{fig:full_deit}.

Once the token selectors are trained for the pairs of layers where \ours will be applied, we freeze the token selectors, modify the second layers in the pairs to be token selective, and finetune the full model on the same training set where it was originally pre-trained (e.g., ImageNet).

%% file: sec/4_experiments.tex
\begin{table}[t!]
\vspace{-5pt}
\footnotesize
  \centering
  \begin{tabular}{c|cc|cc}
    \hline
    \multirow{2}{*}{{Encoder}} & \multicolumn{2}{c|}{\emph{NYUD Depth v2}} & \multicolumn{2}{c}{\emph{KITTI}}\\
    & Abs Rel $\downarrow$ & $\delta\! < \!1.25$ $\uparrow$ & Abs Rel $\downarrow$ & $\delta \! <\! 1.25$ $\uparrow$\\
    \hline
    DeiT-Tiny & 0.101 & 0.896 & 0.063 & 0.956 \\
    + \ours & 0.103 & 0.891 & 0.063 & 0.954\\
    \hline
  \end{tabular}
  \vspace{-7pt}
  \caption{\small Monocular depth estimation on NYU Depth v2 and KITTI using DeiT-Tiny and DeiT-Tiny with \ours as the encoders, respectively. We use NeWCRFs~\cite{yuan2022neural} as the base architecture. }
  \label{tab:nyud}
  \vspace{-9pt}
\end{table}

\section{Experiments}
\label{sec:experiments} \vspace{-3pt}
We evaluate \ours on the ImageNet classification benchmark. We further utilize it for monocular depth estimation and demonstrate its effectiveness. Note that in this short paper, we present representative results to showcase the efficacy of our proposed approach. We plan to include more comprehensive results in the full paper.

\subsection{Experimental Setup}\vspace{-3pt}

\noindent \textbf{Datasets and Evaluation:} We perform standard image classification evaluation on ImageNet~\cite{russakovsky2015imagenet}. We further use classification networks as encoders for monocular depth estimation on NYU Depth v2~\cite{silberman2012indoor} and KITTI~\cite{kitti}, where we use standard depth accuracy metrics such as Absolute Relative Error (Abs Rel) and and $\delta < 1.25$; see~\cite{eigen2014depth} for mathematical definitions of these metrics.




\noindent \textbf{Networks and Implementation:} 
In this work we use DeiT-Tiny~\cite{touvron2021training} as the base model as it is one of the standard vision transformers and the computation resource required for training is more accessible, e.g., we use 2 Nvidia A100 GPUs. For ImageNet evaluation, we train the token selector and finetune DeiT-Tiny with \ours both on ImageNet training data. Since DeiT-Tiny has 12 layers, we apply \ours to the 2nd, 4th, 6th, 8th, and 10th layers, using 80\% tokens for self-attention at each \ours layer. For depth estimation, we use NeWCRFs~\cite{yuan2022neural} as the base model and replace the encoder to be either DeiT-Tiny or DeiT-Tiny with \ours.



\footnotetext{The original token merging method reduces tokens by merging and cannot be used for dense prediction tasks which will require distinct features for all the image patches. \cite{bolya2023token} uses token merging to accelerate stable diffusion, but it requires additional steps to un-merge the tokens. Moreover, while un-merging works for image generation, it may not work for other dense tasks such as depth estimation, since similar-looking patches may have very different depth values.}

\footnotetext{Computation reduction by VTC-LFC also includes channel pruning.}

\subsection{Image Classification}\vspace{-3pt}
Table~\ref{tab:imagenet} summarizes the evaluation results on ImageNet-1K classification benchmark. By applying \ours, we significantly reduce the computation of DeiT-Tiny by nearly $25\%$ while maintaining the classification accuracy. Since the full set of tokens are preserved, our \ours network can be readily used as an encoder for dense prediction tasks. On the other hand, while existing works like ToMe~\cite{bolya2023token} and a-STAR~\cite{zhangsynergistic} can also significantly reduce computation, they rely on reducing the number of tokens, i.e., only a small number of tokens are kept after going through the network. This works for classification but makes it challenging to use such models as encoders for dense tasks. Furthermore, these existing methods result in accuracy drops.

\begin{figure}[t!]
    \vspace{-4pt}
    \centering
    \includegraphics[width=0.32\textwidth]{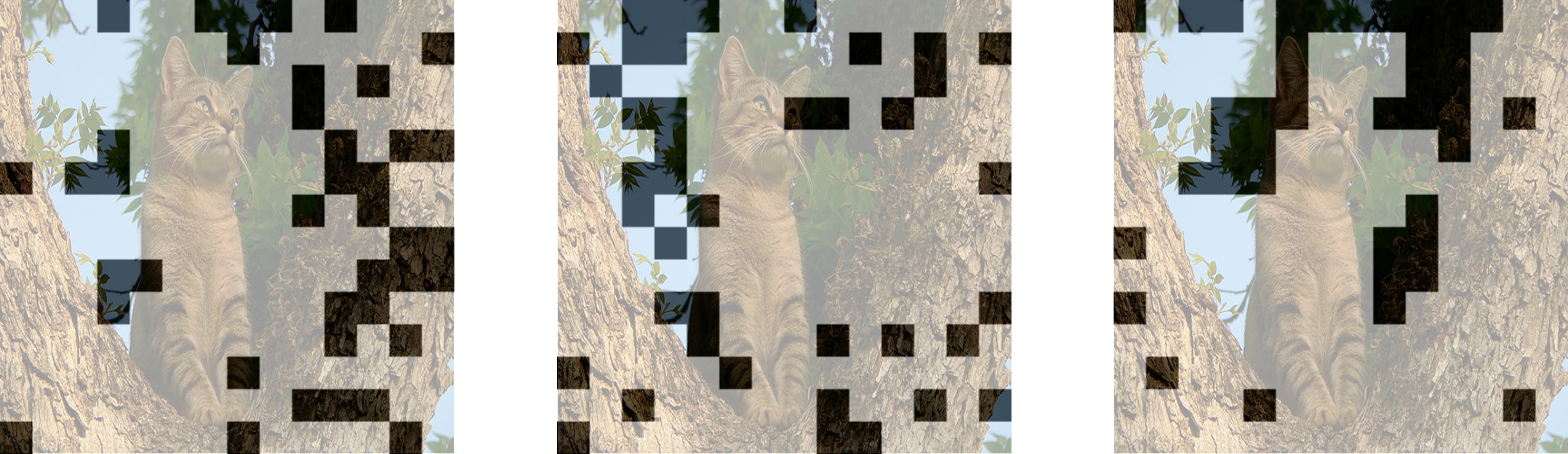}
    \vspace{-8pt}
    \caption{\small Visualization of token selection at the 2nd, 6th, and 10th layers. The input is a cat image with the cat at the center. Dark colors indicate the patches that do not take part in self-attention.}
    \label{fig:vis}
    \vspace{-10pt}
\end{figure}

Fig.~\ref{fig:vis} visualizes the token selections at sample \ours layers (second, sixth, and tenth) of the network. It can be seen that most patches of the main object (cat) are selected for self-attention and the network selects different subsets of background patches at different layers. For instance, at these layers, the network selects background patches from the right, left, and top parts of the image, respectively, while always selecting most of the cat patches.

\vspace{-1pt}
\subsection{Monocular Depth Estimation}\vspace{-3pt}
\ours allows the modified network to be used as an encoder for dense prediction. In Table~\ref{tab:nyud}, we see that our DeiT-Tiny w/ \ours network provides similar depth estimation performance on both NYU Depth v2 and KITTI benchmarks, although it uses considerably fewer tokens for self-attention in several layers, as compared to using the original DeiT-Tiny as an encoder.


%% file: sec/5_conclusion.tex
\section{Conclusion}
\label{sec:conclusion} \vspace{-3pt}
In this paper, we proposed \ours, a novel token selective attention approach to make vision transformers more efficient. Based on predicted attention maps, a subset of important tokens are selected for self-attention while the rest bypass the transformer layer. Unlike existing works, \ours maintains the full set of tokens, making the network readily usable for dense prediction. Our evaluation results indicate that \ours can significantly reduce computation costs while maintaining accuracy on both classification and dense prediction tasks.